\documentclass{article}
\usepackage{ijcai20}

\usepackage{times}
\usepackage{soul}
\usepackage{url}
\usepackage[hidelinks]{hyperref}
\usepackage[utf8]{inputenc}
\usepackage[small]{caption}
\usepackage{graphicx}
\usepackage{amsmath}
\usepackage{amsthm}
\usepackage{booktabs}
\usepackage{algorithm}
\usepackage{algorithmic}
\usepackage{xcolor}
\usepackage{amsmath,amssymb}
\usepackage{easyeqn}
\usepackage[symbol]{footmisc}

\title{Linking Bank Clients using Graph Neural Networks Powered by Rich Transactional Data}

\author{
Valentina Shumovskaia$^{2,}$\footnote{Equal contribution}\and
Kirill Fedyanin$^{1, *}$\and
Ivan Sukharev$^2$\and
Dmitry Berestnev$^2$\And
Maxim Panov$^1$
\affiliations
$^1$CDISE, Skolkovo Institute of Science and Technology (Skoltech) \\
$^2$Risk Modeling \& Research, Sberbank
\emails
\{k.fedyanin, m.panov\}@skoltech.ru,
\{VSShumovskaya, Sukharev.I.Vi, DABerestnev\}@sberbank.ru
}

\hypersetup{draft}

\begin{document}

\maketitle

\begin{abstract}
  Financial institutions obtain enormous amounts of data about user transactions and money transfers, which can be considered as a large graph dynamically changing in time. In this work, we focus on the task of predicting new interactions in the network of bank clients and treat it as a link prediction problem. We propose a new graph neural network model, which uses not only the topological structure of the network but rich time-series data available for the graph nodes and edges. We evaluate the developed method using the data provided by a large European bank for several years. The proposed model outperforms the existing approaches, including other neural network models, with a significant gap in ROC AUC score on link prediction problem and also allows to improve the quality of credit scoring.
\end{abstract}

\section{Introduction}
\label{sec:intro}

  It is important for the financial institutions to know their client well in order to mitigate credit risks~\cite{Siddiqi2012}, deal with fraud~\cite{Phua2010} and recommend relevant services~\cite{Bruss2019}. One of the defining properties of a particular bank client is his or her social and financial interactions with other people. It motivates to look on the bank clients as on the network of interconnected agents~\cite{Tran2019,Bruss2019,Weber2018}. Thus, graph-based approaches can help to leverage this kind of data and solve the above mentioned problems more efficiently.

  Importantly, information about clients and especially about their neighborhood is never complete -- market is competitive and we can not expect all the people to use the same bank. Thus, some of the financial interactions are effectively hidden from the bank. That leads to the necessity to uncover hidden connections between clients with limited amount of information which can be done using link prediction approaches~\cite{Wang2015}.

  From the other hand, the financial networks have two notable features. First one is the size -- the number of clients can be of order of millions and the number of transactions is estimated in billions. The second important feature is the dynamic structure of considered networks -- the neighborhood of each client is ever-evolving. The classical link prediction algorithms are only capable of working with graphs of a much smaller size, while the temporal component is usually not considered~\cite{Wang2015}. Recently, several studies addressed large-scale graphs~\cite{Ying2018} as well as temporal networks~\cite{Pareja2019}. However, only few works consider the financial networks, see, for example, \cite{Bruss2019} and~\cite{Tran2019}.

  We base our research on the well developed paradigms of graph mining with neural networks including graph convolution networks~\cite{Kipf2016SemiSupervisedCW,Hamilton2017InductiveRL}, graph attention networks~\cite{Velickovic2017GraphAN} and SEAL framework for link prediction~\cite{zhang2018link}. The considered approaches consistently show state-of-the-art results in many applications but, to the best of our knowledge, were not yet used for financial networks. Our key contributions can be formulated as follows:
  \begin{itemize}
    \item We build a scalable approach to link prediction in temporal graphs with the focus on extensive usage of Recursive Neural Networks (RNNs) both as feature generators for graph nodes and as a trainable attention mechanism for the graph edges.

    \item We propose several modifications to graph pooling procedures including the usage of two node convolutions instead of sortpooling~\cite{zhang2018link} and neighborhood prioritization by Weisfeiler-Lehman labeling~\cite{zhang2017weisfeiler}. 

    \item We validate the proposed approaches on the link prediction and credit scoring problems for the real-world financial network with millions of nodes and billions of edges. Our experiments show that our improved models perform significantly better than the standard ones and efficiently exploit rich transactional data available for the edges and nodes while allowing to proceed large-scale graphs.
  \end{itemize}


\section{Problem and Data}
\label{sec:problem}
  From the prospectives of network science and data analysis, the considered problem of linking bank clients is the link prediction problem in graphs with two notable peculiarities. The first one is that the considered graph of clients and transactions between them is very large, having the order of millions of nodes and billions of edges. The second peculiarity is that both nodes and edges have rather complex attributes represented by times series of bank transactions of different types. We want to note that such kind of problem is not limited to banking as graphs with similar structure appear in social networks, telecom companies, and other scenarios, where we consider some objects as nodes and a certain type of communication between them. Thus, the algorithms developed in our work might be applicable beyond banking for any link prediction problem with times-series attributes.

  In what follows, we first discuss the dataset studied in our work and then explain some peculiarities of the problem statement.

\subsection{Dataset}
  The considered dataset is obtained from one of the large European banks. The data consists of user transactions and money transfers between users during five years. All the data is depersonalized with each transaction being described by timestamp, amount and currency.
  Thus, we observe a graph $G(V, E)$ with a set of vertices $V$ and a set of edges $E$. Here, an edge $(i, j) \in E$ means that there was at least one transfer between a pair of clients $i$ and $j$ over the observed time period. Each node $i \in V $ is represented by a time series of transactions for client $i$, while each edge $(i, j) \in E$ is represented by a time series of transfers between clients $i$ and $j$. Finally, we obtain a huge 86-million nodes graph with about 4 billions of edges.

  Such graph size makes its analysis difficult to approach, since the majority of the graph processing methods aimed to solve node classification, graph classification or link prediction problems are suitable for graphs of much smaller size~~\cite{Hamilton2017}. The time complexity of such methods usually grows at least as $n^2$, where $n$ is number of nodes, limiting the possible graph sizes to several thousands of nodes and up to a one hundred thousand of edges.

  As a result, when we work with a particular node or with a particular edge, we are forced to consider certain subgraphs around the target node or the target pair of nodes (for example, see~\cite{zhang2017weisfeiler}). In this work, we follow this approach and consider the subgraph around target nodes extracting hop 1 or hop 2 neighbors.

\subsection{Problem Statement and Validation}
\label{sec:validation}
  Our goal is to determine how stable is the relationship between nodes. We start by describing out-of-time validation (see a similar approach in~\cite{Liben-nowell07thelink-prediction}), more specifically, we consider time interval time $[t_0, t_1]$ for $t_0 < t_1$ and use all the information available (e.g. all the transactions and transfers) as an information encoded in a graph. Given the information available for the period $[t_0, t_1]$ we aim to predict the structure of the graph for the time interval $[t_1, t_2]$ with $t_1 < t_2$. In what follows, we say that there is an edge between two nodes in a graph for the certain time period if there was at least one transaction between these nodes during the considered period. Thus, we end up with link prediction problem where the pair of nodes is described by the graph structure and attributes during the period $[t_0, t_1]$ and the target label corresponds to the existence of the transaction between the pair of nodes during the period $[t_1, t_2]$. In all the experiments below we take $t_1 - t_0$ equal to one year and $t_2 - t_1$ equal to 3 months.

  We note that usually link prediction models are validated in a different way, e.g. by edge sampling~\cite{Liben-nowell07thelink-prediction}.  In this approach, the whole set $E$ is considered as positive samples, while negative samples are constructed by taking $\alpha |E|$ node pairs ($\alpha > 0$ is a hyperparameter and  $|\cdot|$ means the set size), which do not intersect with $E$. Then, the subgraph is passed to the link prediction algorithm, hiding the link, if it exists. In order to build training, validation and test parts, one divides positive and negative edge sets into three corresponding non-intersecting sets.

  However, we think that for the time-evolving graphs in general and banking data in particular the out-of-time validation is more sensible. Thus, in this work, we focus on the out-of-time validation, while still providing a part of the experiments for both settings.


\section{Neural Network Model for Link Prediction with Transactional Data}
\label{sec:method}
  In this section, we describe the proposed neural network for solving a link prediction task powered by rich transactional data. The most challenging part is the work with transactional data itself, which is basically a multidimensional time series.

\begin{figure*}[ht!]
    \centering
    \includegraphics[scale=0.12]{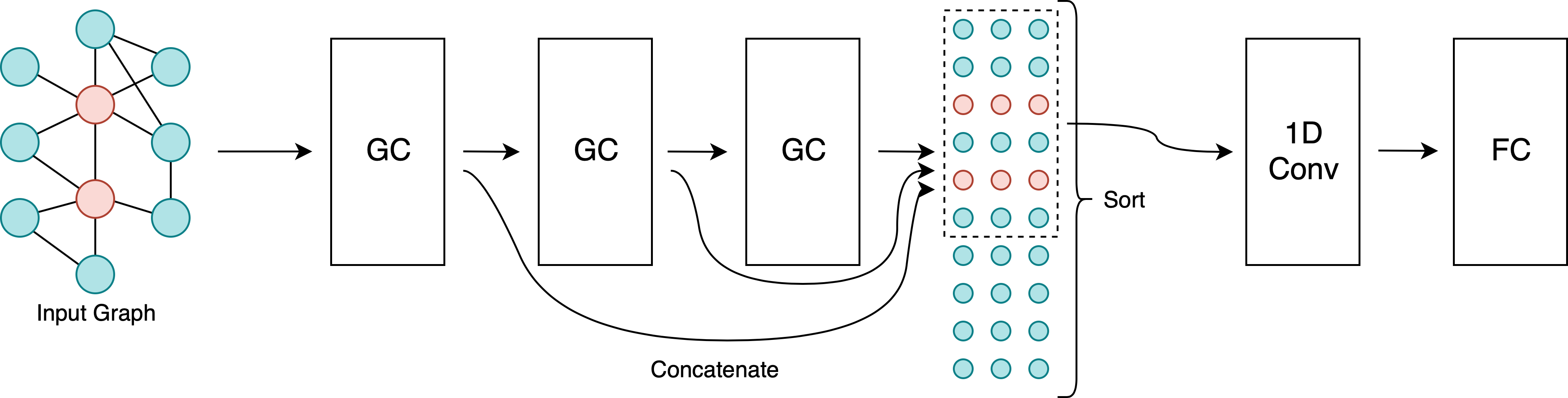}
    \caption{SEAL architecture. The input graph is passed to a series of Graph Convolution (GC) layers. The obtained nodes features are sorted and pooled with SortPooling layer, then they are passed to 1-D Convolution layer (1D Conv) and Fully Connected (FC) layer.}
    \label{fig:seal}
  \end{figure*}

  As a base graph neural network, we take SEAL framework~\cite{zhang2018link}. Its input parameters are an adjacency matrix of a graph $A \in \mathbb{R}^{n \times n}$ and a node feature matrix $X \in \mathbb{R}^{n \times d}$ with each row containing a feature vector for the corresponding node. Then SEAL considers the neigborhood subgraph for the target pair of nodes and performs several graph convolutions followed by sortpooling operation and fully connected layers, see Figure~\ref{fig:seal}. However, the considered network of bank transactions does not have an explicit adjacency matrix or a vector of features as both clients and interactions between them are represented by time series. In the following, we are going to adapt SEAL framework to work with time series data by processing them with RNN. Moreover, we make a number of specific improvements to the structure of SEAL model making it more efficient.
   
\subsection{Recursive Neural Network Powers Graph Neural Network}
\label{sec:rnn}
\subsubsection{RNN as Feature Generator}
\label{sec:feature_generation}
  The powerful way of working with time series data is to build a Recurrent Neural Network (RNN, \cite{connor1994recurrent}). The main question is what objective function RNN should target. We suggest to pretrain RNN model on the credit scoring problem similar to~\cite{Babaev2019}, see also additional details in Section~\ref{sec:credit_scoring}. The model takes a time-series of user transactions and aims to predict the credit default.  For that purpose, we take a quiet simple Recurrent Neural Network, which consists of GRU cell~\cite{cho2014learning}, followed by a series of fully connected layers. Importantly, such RNN model learns in the intermediate layers the meaningful vector representation for the transactions of the client. In the following, we call these vectors \textit{embedded transactions} and use them as node feature vectors $X$ in all the considered graph neural network models.

\subsubsection{RNN as Attention Mechanism}
\label{sec:attention}
  The question of processing time series corresponding to the graph edges is even more challenging than the one for nodes. The simplest way is just to ignore the whole time series and consider binary adjacency matrix with edges present for pairs of nodes with at least one transfer between them. However, in this case we lose significant amount of important information as the properties of transfers between clients are apparently directly linked with our link prediction objective.

  \begin{figure}[t!]
    \centering
    \includegraphics[scale=0.105]{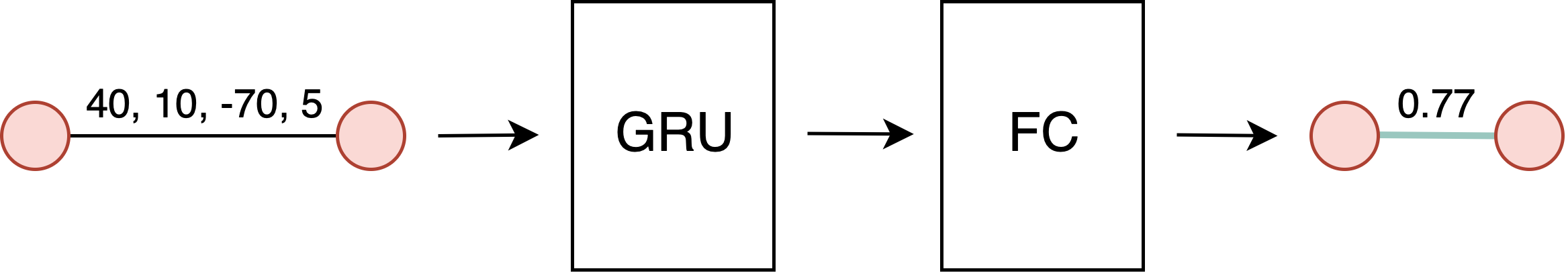}
    \caption{RNN for link prediction architecture.}
  \label{fig:rnn}
  \end{figure}

  In order to get the full use of the data, we first note that one can consider a RNN model predicting the link  between two nodes using solely the time series of transfers between them, see Figure~\ref{fig:rnn}. However, such a RNN model does not allow us to detect \textit{new} possible connections since there is no data about interaction between users in this case. To overcome that drawback, a model based on a transactional graph can be used.

  We first note that standard graph convolutional architectures (like GCN~\cite{Kipf2016SemiSupervisedCW} or SEAL~\cite{zhang2018link}) perform convolution operation by simple averaging over the neighborhood:
  \begin{EQA}[c]
    h'_i = \sigma \Big(\frac{1}{|\mathcal{N}_i|} \sum_{j \in \mathcal{N}_i} W h_j\Big), i = 1, \dots, n,
  \end{EQA}
  where $(h_1, \dots, h_n)$ are node embedding vectors before the convolution operation, $(h'_1, \dots, h'_n)$ are their counterparts after it, $W$ are learnable weights, $\mathcal N_i$ is a set of immediate neighbors of node $i$ and, finally, $\sigma$ is an activation function. The averaging operation implies that all the neigbors have an equal influence on the considered nodes which is apparently very unnatural in the majority of applications. 

  Graph Attention Networks~\cite{Velickovic2017GraphAN} mitigate this problem by introducing weights $\alpha_{ij}$ and consider the weighted sum:
  \begin{EQA}[c]
    h'_i = \sigma \Big(\frac{1}{|\mathcal{N}_i|} \sum_{j \in \mathcal{N}_i} \alpha_{ij} W h_j\Big).
  \end{EQA}
  However, in the work~\cite{Velickovic2017GraphAN} coefficients $\alpha_{ij}$ are computed solely basing on node features $X$. Instead, in order to use the full information about the graph, we propose to use the probabilities of the links between nodes output by RNN model as weights in the adjacency matrix, which then is passed to graph neural network.

\begin{figure*}[ht!]
    \centering
    \includegraphics[scale=0.12]{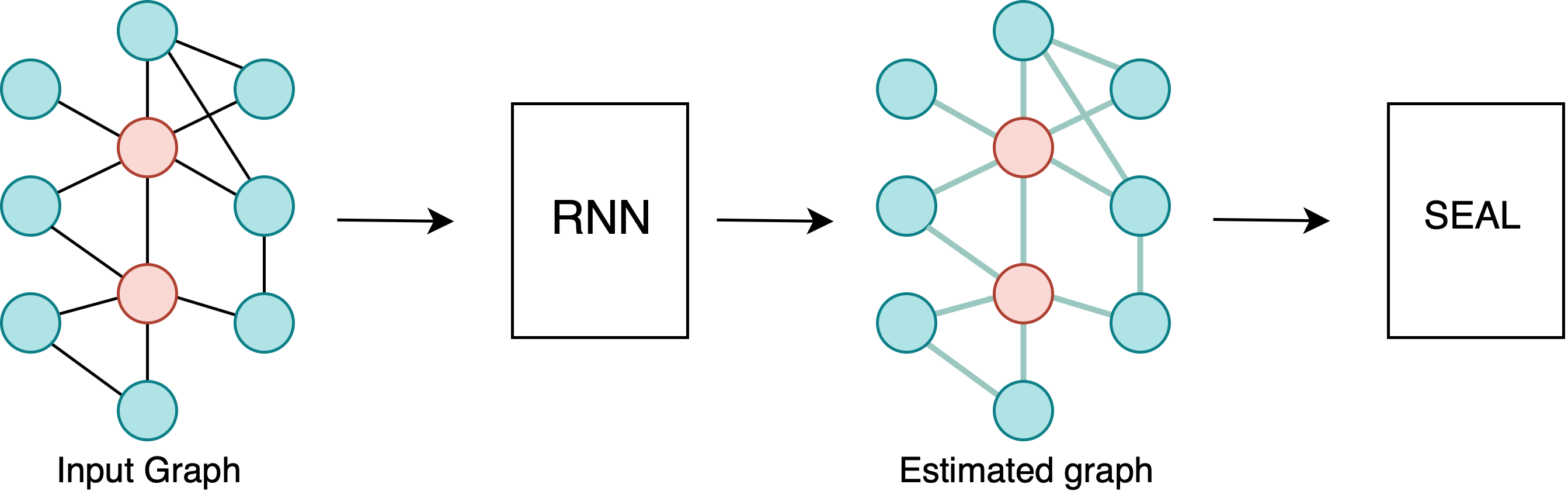}
    \caption{SEAL-RNN model architecture. After extracting an enclosing subgraph around the target link, all pairs in that subgraph are estimated by RNN, so we get a weighted adjacency matrix. Weighted adjacency matrix $A$ and generated nodes features $X$ are passed into SEAL model.}
    \label{fig:seal_rnn}
  \end{figure*}

  The resulting model is called SEAL-RNN, see the architecture on Figure~\ref{fig:seal_rnn}. After extracting an enclosing subgraph around the target link, all time series corresponding to edges are processed by RNN and the output probabilities are used to form weighted adjacency matrix $\tilde{A}$ which together with generated nodes features $X$ are passed into SEAL model.

\subsection{Graph Neural Network (2-SEAL)}

\subsubsection{Pooling}
\label{sec:pooling}
\begin{figure*}[ht!]
    \centering
    \includegraphics[scale=0.12]{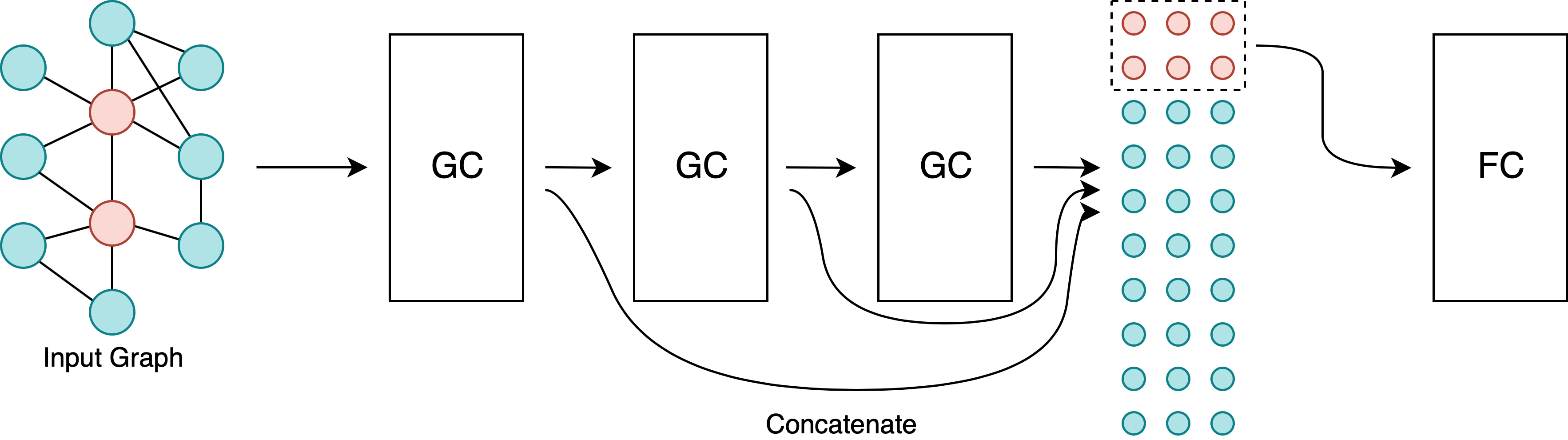}
    \caption{2-SEAL architecture. At first, the input graph is passed to a series of Graph Convolution (GC) layers. Then the obtained nodes features of two target nodes (between which the link is predicted) are passed to Fully Connected (FC) layer.}
    \label{fig:2seal}
  \end{figure*}

  We propose another pooling operation instead of sortpooling in the SEAL model. Sortpooling layer holds $K$ (hyperparameter) most valuable in the sense of sorting (descending order) node embeddings while filtering out the other embeddings. In contrast, we suggest taking embeddings of two nodes, between which we aim to predict the link. The idea is natural since we want to predict the link between exactly these two nodes, while their  embeddings still contain information about the neighboring nodes. Most importantly, it reduces the number of learned parameters in the neural network, and we do not need neither a sorting operation, nor 1-D convolution after pooling (the purpose of 1-D convolution in SEAL framework is to reduce the size of obtained output, which is $K \times d$, where $d$ is a sum of node features dimension and dimensions of the graph convolution outputs). We name the proposed model 2-SEAL, see the schematic representation on Figure~\ref{fig:2seal}. 

\subsubsection{Modified Structural Labels}
\label{sec:struct_labels}
  Working in terms of out-of-time validation, we decided to change structural labels proposed in the SEAL framework. In SEAL framework, each node receives a structural label generated by a Double-Radius-Node-Labelling procedure, which meets the following conditions:
  \begin{enumerate}
    \item two target nodes $x$ and $y$ have label `1'; 

    \item nodes with different distances to both $x$ and $y$ have different labels.
  \end{enumerate}
  The aim of the labels is to encode some of the topological information about the graph structure.
  These structural labels are concatenated with initial node features (if exist), and passed to neural network as node features. The labelling function ($i$ is a node index) is the following:
  \begin{EQA}[c]
    f(i) = 1 + \min(d_x, d_y) + (d/2) \bigl[(d/2) + (d\%2) - 1\bigr],
  \end{EQA}
  where $d_x = d(i, x)$, $d_y = d(i, y)$, $d = d_x + d_y$, $(d / 2)$ and $d\%2$ are the integer quotient and remainder of division respectively, while $d(\cdot, \cdot)$ is distance between nodes. Authors of initial paper suggest to take into account all subgraph nodes except $y$ during computing distance $d_x$, and similarly for $d_y$. 
  
  We suggest not to hide nodes $y$ and $x$ during finding distances $d_x, d_y$. That better suits out-of-time validation by allowing to keep in data patterns for all kinds of combinations of link existence in the observed graph and link existence in the future. 

\section{Related Work}
\label{sec:related_work}
  The idea to consider bank clients as a large network of interconnected agents was raised in the past several years~\cite{Tran2019,Bruss2019,Weber2018}. The number of bank clients counts in millions, so we solve the link prediction problem for graphs with millions of nodes, which requires the usage of scalable methods. There are few ways to handle the graphs of such size mentioned in the literature, mostly being the simple heuristics that compute some statistics for the immediate neighborhoods of target nodes, for example, Common Neighbors~\cite{Newman2001ClusteringAP}, Adamic-Adar~\cite{Adamic2001FriendsAN} and others~\cite{Wang2015}. However, these models are not trainable and do not use the information about node features, which limits their performance in real-world applications.

  The main challenge in the construction of machine learning models for link prediction is to handle variation in the graph size. One approach is presented in WLNM~\cite{zhang2017weisfeiler} -- it is to use Weisfeiler-Lehman structural labels~\cite{Weisfeiler1968ReductionOA} to prioritize nodes and to leave only the important one from the immediate neighborhood of evaluated nodes. After that, we can use regular dense-connected neural networks.

  The graph convolution networks~\cite{Kipf2016SemiSupervisedCW} showed good performance on graph datasets. Original GCN is supposed to use the whole graph, and it is prohibitive for the graph on a scale of millions of nodes. In~\cite{zhang2018link}, the SEAL framework was proposed, which is to extract enclosing subgraphs around the target link and include such a pooling layer in the neural network architecture, which holds the fixed number of nodes for every subgraph. This allows using the model on arbitrarily sized graphs.

  The novel Graph attention model GAT~\cite{Velickovic2017GraphAN} allows specifying different weights to different nodes in the neighborhoods. That approach could be used to leverage sequence information on the edges by adding attention coefficients to the graph convolutions.

\section{Experiments}
\label{sec:experiments}

\subsection{Dataset Preprocessing}
\label{sec:preprocessing}
\begin{figure}[t!]
    \centering
    \includegraphics[scale=0.05]{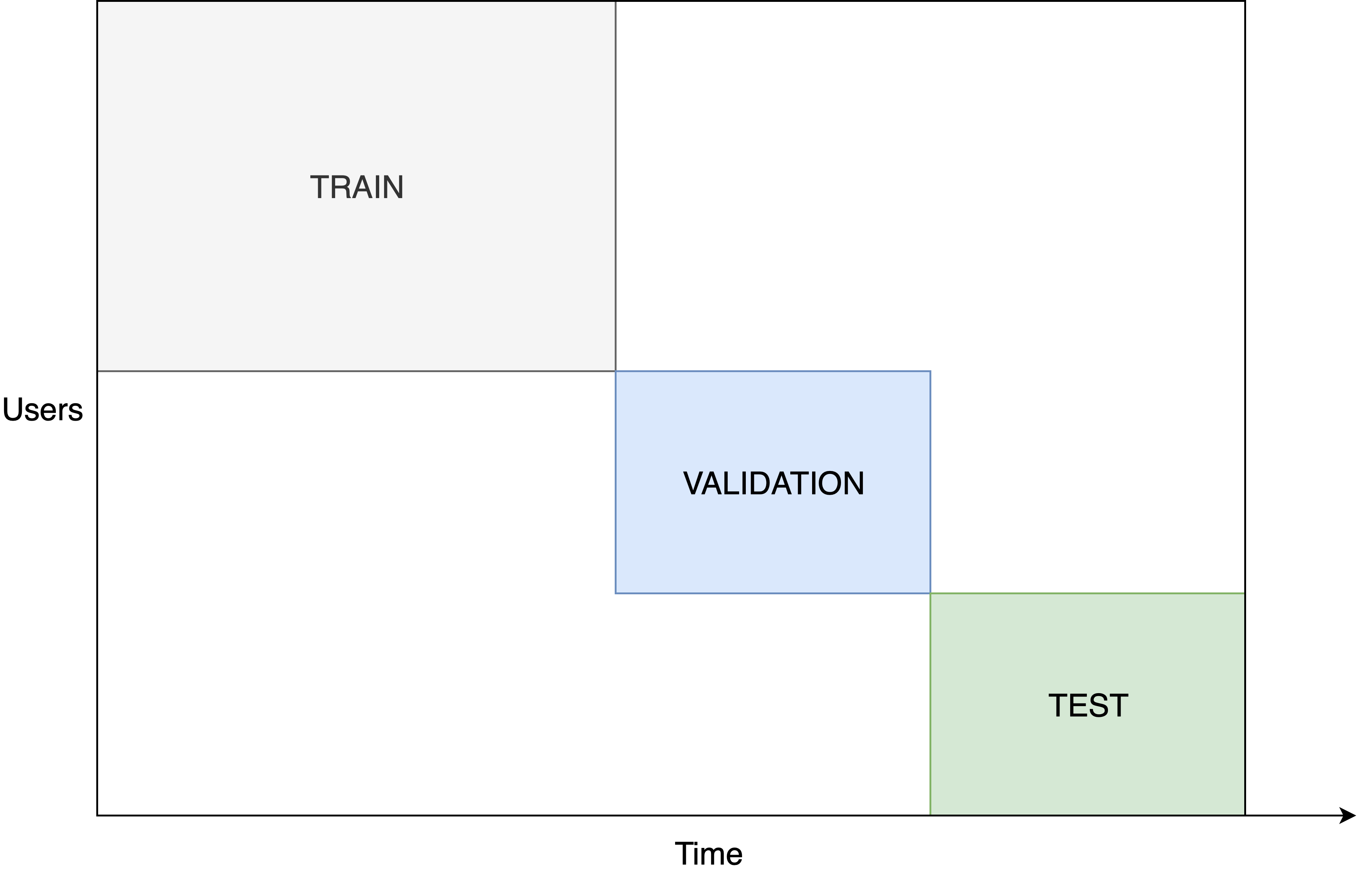}
    \caption{Data split for train, validation and test.}
    \label{fig:split}
  \end{figure}

  Firstly, we divide the whole time interval and the set of user IDs into three non-intersecting parts: first three years, fourth year, and fifth year, they correspond to training, validation, and test time and users segments, see Figure~\ref{fig:split}. Taking a point in one of the time intervals, we define the base and the target segment. The base segment corresponds to the time segment before the point, while the target segment corresponds to time after. For edge sampling validation, we observe graph state restricted to the base segment, while the target is whether there was at least one transfer between users during this time. For the out-of-time validation setting, the target is whether there is at least one transfer between users during the target segment.
  We consider ROC-AUC measure as a quality metric for the link prediction task.

\subsection{Baselines}
\label{sec:baselines}
  Due to the need in the scalability we consider only simple similarity-based approaches, such as Common Neighbors, Adamic-Adar Index, Resource Allocation, Jaccard Index and Preferential Attachment, as baselines for our task (see~\cite{Wang2015} for the description of the methods). Also, we take the SEAL model~\cite{zhang2018link} as a baseline (with embedded transactions concatenated with structural labels as node features). The results can be found in Table~\ref{tab:heuristics}. As we may see, the results obtained from simple heuristic methods are beaten by the neural network solution. Also, there is a gap in the ROC AUC score for different validation settings. It could be explained by the fact that the problem of prediction into the future is a more difficult problem than finding hidden links in the current graph state.

\subsection{Implementation details}
\label{sec:implementation}
  We use PyTorch~\cite{PyTorch} and PyTorch Geometric~\cite{Fey/Lenssen/2019} to implement the models. Each model was trained with Adam optimizer~\cite{Kingma2014AdamAM} using learning rate scheduler and hyperparameter optimization~\cite{Bergstra2013MakingAS} for the number of layers, size of the layers and initial learning rate. We used the server with single GPU (NVIDIA Tesla P100), 32 CPU cores Intel i7 and 512 GB of RAM in all the experiments.

\subsection{Link Prediction Results}
\label{sec:results}
  \begin{table}
    \centering
    \begin{tabular}{lrr}  
      \toprule
      Method  & Edge sampling & Out of time \\
      \midrule
      Common Neighbors        & 0.398 &  0.629 \\
      Adamic-Adar             & 0.391 & \textbf{0.646} \\
      Resource Allocation     & 0.35  & 0.639 \\
      Jacard Index            & 0.284 & 0.62  \\
      Preferential Attachment & \textbf{0.746} & 0.497 \\
      SEAL                    & \textbf{0.85}  & \textbf{0.77} \\
      \bottomrule
    \end{tabular}
    \caption{Heuristics approaches and SEAL (with embedded transactions and structural labels as node features) results for banking data (ROC AUC).}
  \label{tab:heuristics}
  \end{table}

  \begin{table}
    \centering
    \begin{tabular}{lrr}  
      \toprule
      Method  & Edge sampling & Out of time \\
      \midrule
      SEAL                    & 0.85   & 0.77 \\
      WL-SEAL                 & 0.87   & 0.75 \\
      2-SEAL                  & \textbf{0.89} & \textbf{0.78} \\
      \bottomrule
    \end{tabular}
    \caption{SEAL (with embedded transactions and structural labels as node features) pooling modifications results for banking data (ROC AUC).}
  \label{tab:pooling}
  \end{table}

  The first improvement of the initial SEAL model is the new pooling operation. SEAL and 2-SEAL models are described in the previous sections (see Sections~\ref{sec:method} and~\ref{sec:related_work}). We additionally consider WL-SEAL pooling operation which is based on the idea of the Weisfeiler-Lehman graph isomorphism test. Quiet similarly to the idea described in~\cite{zhang2017weisfeiler}, we propose to color nodes of enclosing subgraphs by the Palette-WL algorithm (Algorithm 3 in~\cite{zhang2017weisfeiler}), thereby get nodes ordering. After that, we take only $K$ (hyperparameter) the most significant nodes of the subgraph, as an input of the neural network. Thus, all subgraphs have the same size, so there is no need for a pooling operation after convolution layers.
  We expect that such pooling is more meaningful in the sense of intuition, but the drawback of such a model is computationally expensiveness of coloring algorithm ($O\bigl(e^{\sqrt{n \log n}}\bigr)$). The results can be found in Table~\ref{tab:pooling}. We observe that both WL-SEAL and 2-SEAL are superior to SEAL. However, 2-SEAL shows the best results, being less computationally expensive model which motivates us to focus the further studies on this model.


  Another set of experiments is devoted to the exploration of the features. In the previous set of experiments on neural networks, we used a concatenation of embedded transactions (the output of an intermediate level of RNN which solves a credit scoring task) and structural labels as node features. We provide experiments in different settings of node features embedded transactions, embedded transactions concatenated with structural labels, structural labels, and modified structural labels (structural labels and modified structural labels are described in Section~\ref{sec:pooling}). Surprisingly, the usage of embedded transactions plays a negative role in the link prediction task. We explain it by the fact that similar purchases do not play a significant role in problems of finding new connections in the network, while network structure and people's connections are a way more important. Also, modified structural labels (without hiding the link) gave us a better performance. 

  The final set of experiments is based on the work with data corresponding to edges (see Table~\ref{tab:rnn_features}), where we consider different RNN-based models, see details in Section~\ref{sec:rnn}. We see that in every setting (except embedded transactions + structural labels for 2-SEAL model), we have a large increase in the ROC AUC score (almost 0.1 in some of the cases) for the proposed models. We conclude that  2-SEAL model with RNN attention is the best model for link prediction for the considered banking dataset.

  The summary of the results can be found in Table~\ref{tab:final}. We observe the significant improvement in the ROC AUC score for the proposed 2-SEAL-RNN model compared to the best heuristic approach and SEAL. 

  \begin{table}
    \centering
    \begin{tabular}{lrrrr}  
      \toprule
      Method  & ET & ET+SL & SL & Modified SL \\
      \midrule
      SEAL                & 0.62 & 0.747 & 0.74 & 0.76 \\ 
      SEAL-RNN            & 0.61 & 0.787 & 0.78 & 0.794 \\
      2-SEAL              & 0.7  & 0.739 & 0.77 & 0.787 \\
      2-SEAL-RNN          & \textbf{0.727} & \textbf{0.804} & \textbf{0.83} & \textbf{0.858}\\
      \bottomrule
    \end{tabular}
    \caption{SEAL pooling modifications results for banking data with embedded transactions (ET) and structural labels (SL) as node features (ROC AUC).}
  \label{tab:rnn_features}
  \end{table}

  \begin{table}
    \centering
    \begin{tabular}{lr}  
      \toprule
      Method  & Result, ROC AUC \\
      \midrule
      Best heuristic  &  0.646 \\ 
      SEAL            & 0.74 \\
      2-SEAL          & 0.79 \\
      2-SEAL-RNN      & \textbf{0.858} \\
      \bottomrule
    \end{tabular}
    \caption{Final results on banking data in out-of-time validation setting.}
  \label{tab:final}
  \end{table}

\subsection{Credit Scoring Results}
\label{sec:credit_scoring}
  In this section, we want to show the applicability of the developed link prediction models to other problems relevant for the banking. One of the most important problems in the bank is to control the risks related to working with clients, especially in the process of issuing a loan.  This problem is called credit scoring~\cite{Siddiqi2012}, and usually the ensemble of predictive models is used, which in particular are based on user transactional data. For example, the RNN model run on time series of transactions has been shown to be very efficient in credit scoring~\cite{Babaev2019}.

  The usage of information available in the network of clients may further improve the prediction quality. We consider the credit scoring dataset of approximately one hundred thousand clients which is a part of our initial dataset. Our experiments show the standard Graph Convolutional Network (GCN)~\cite{Kipf2016SemiSupervisedCW} trained on these data improves over baseline RNN model by $0.8 \%$ Gini, see Table~\ref{tab:scoring}. However, GCN model is known to treat all the neighboring nodes equally without any prioritization (see discussion in Section~\ref{sec:rnn}), which is apparently not correct for the bank clients some of which have much more influence on the particular client than the others. This issue was addressed in the literature by introducing graph attention mechanism based on the available node features~\cite{Velickovic2017GraphAN}.

  In our work, we propose to use the developed link prediction model (2-SEAL-RNN) as an attention mechanism by reweighing the neighbouring nodes with coefficients proportional to the probabilities of the connection output by the link prediction model. Unlike standard Graph Attention Networks~\cite{Velickovic2017GraphAN}, our attention mechanism considers not only node features but also the topology of the graph while still allowing to train the final credit scoring model in end-to-end fashion. In Table~\ref{tab:scoring}, we compare GCN performance which use binary adjacency matrices, and adjacency matrices weighed by the link prediction model. We note that we use the embeddings obtained by RNN as node features in both models. The results show that the link prediction model used as an attention in GCN allows almost to double the effect of considering graph structure in credit scoring problem. We believe that the further study of the link prediction based attentions in graph neural network may lead to even better credit scoring models.

  \begin{table}
  \centering
  \begin{tabular}{lc}  
    \toprule
    Method & Result, $\Delta$ in Gini index \\
    \midrule
    Standard GCN & + 0.8\% \\
    GCN with LP-based attention & + 1.4\% \\
    \bottomrule
  \end{tabular}
  \caption{Gini index scores for GNNs models applied to credit scoring task in comparison with results obtained by RNN run on transactional data for each user.}
  \label{tab:scoring}
\end{table}

\section{Conclusion}
\label{sec:conclusions}
  In this work, we developed the graph convolutional neural network, which can efficiently solve the link prediction problem in large-scale temporal graphs appearing in banking data. Our study shows that to benefit from the rich transaction data fully, one needs to efficiently represent such data and carefully design the structure of the neural network. Importantly, we show the effectiveness of Recursive Neural Networks as building blocks of temporal graph neural network, including a non-standard approach to the construction of attention mechanism based on RNNs. We also modify the existing GNN pooling procedures to simplify and robustify them. The developed models significantly improve over baselines and provide high-quality predictions on the existence of stable links between clients, which enables bank with a powerful instrument for the analysis of clients' network. In particular, we show that the usage of the obtained link prediction model as an attention module in the graph convolutional neural network allows to improve the quality of credit scoring.

\bibliographystyle{named}
\bibliography{lp_rtd}

\end{document}